\documentclass[10pt,journal]{IEEEtran}
\usepackage{amsmath,amsfonts}
\usepackage{algorithmic}
\usepackage{array}
\usepackage{pifont}
\usepackage{hyperref}
\usepackage{subcaption}
\usepackage{textcomp}
\usepackage{stfloats}
\usepackage{url}
\usepackage{verbatim}
\usepackage[table]{xcolor}
\usepackage{graphicx}
\usepackage{multirow} 
\def\BibTeX{{\rm B\kern-.05em{\sc i\kern-.025em b}\kern-.08em
    T\kern-.1667em\lower.7ex\hbox{E}\kern-.125emX}}
\usepackage{balance}\usepackage{svg}
\UseRawInputEncoding{}
\usepackage{cite}
\usepackage{amsmath,amssymb,amsfonts}
\usepackage{algorithmic}
\usepackage{textcomp}
\usepackage{xcolor}
\usepackage{url}
\def\BibTeX{{\rm B\kern-.05em{\sc i\kern-.025em b}\kern-.08em
    T\kern-.1667em\lower.7ex\hbox{E}\kern-.125emX}}
    
\begin{document}

\title{
MVTD: A Benchmark Dataset for Maritime Visual Object Tracking
}
\author{Ahsan Baidar Bakht, Muhayy Ud Din, Sajid Javed, Irfan Hussain
\thanks{$^{1}$ Khalifa University Center for Autonomous Robotic Systems (KUCARS), Khalifa University, United Arab Emirates.}%
 \thanks{$^{*}$This work is supported by the Khalifa University of Science and Technology under Award No. 8434000534, CIRA-2021-085,
RC1-2018-KUCARS, KU-Stanford :8474000605.}%
\thanks{$^{*}$ Corresponding Author, Email: irfan.hussain@ku.ac.ae}
}

\maketitle

\begin{abstract}

Visual Object Tracking (VOT) is a fundamental task with widespread applications in autonomous navigation, surveillance, and maritime robotics. Despite significant advances in generic object tracking, maritime environments continue to present unique challenges, including specular water reflections, low-contrast targets, dynamically changing backgrounds, and frequent occlusions. These complexities significantly degrade the performance of state-of-the-art tracking algorithms, highlighting the need for domain-specific datasets.
To address this gap, we introduce the Maritime Visual Tracking Dataset (MVTD), a comprehensive and publicly available benchmark specifically designed for maritime VOT. MVTD comprises 182 high-resolution video sequences, totaling approximately 150,000 frames, and includes four representative object classes: boat, ship, sailboat, and unmanned surface vehicle (USV). The dataset captures a diverse range of operational conditions and maritime scenarios, reflecting the real-world complexities of maritime environments.
We evaluated 14 recent SOTA tracking algorithms on the MVTD benchmark and observed substantial performance degradation compared to their performance on general-purpose datasets. However, when fine-tuned on MVTD, these models demonstrate significant performance gains, underscoring the effectiveness of domain adaptation and the importance of transfer learning in specialized tracking contexts.
The MVTD dataset fills a critical gap in the visual tracking community by providing a realistic and challenging benchmark for maritime scenarios. Dataset and Source Code can be accessed here "https://github.com/AhsanBaidar/MVTD"

\end{abstract}

\section{Introduction}

Visual Object Tracking (VOT) is a fundamental task and perhaps an open problem in computer vision \cite{Survey1}.
Given the initial location of a target object in the first frame, typically defined by a bounding box or segmentation mask, the goal of VOT is to accurately and consistently estimate the target's state across subsequent video frames \cite{Survey2}.
VOT has a wide range of applications, including autonomous navigation \cite{Autonomous_Navigation,Waseem3}, video surveillance \cite{Surveillance}, pose estimation \cite{pose_estimation}, sports video analysis \cite{Sport_Analysis}, medical video analysis \cite{Medical_Tracking}, underwater monitoring \cite{Waseem1, Waseem6, Waseem4} and human behavior understanding \cite{behaviour_analysis}.
Despite substantial progress in recent years, VOT continues to pose significant challenges due to factors such as occlusion, scale variation, illumination changes, background clutter, and complex object motion \cite{Survey}. These challenges, however, also present exciting opportunities for advancing the frontiers of learning-based tracking, inspiring novel algorithmic frameworks and robust representation learning techniques.

\begin{figure}[t]
  \centering
  \includegraphics[width=\linewidth]{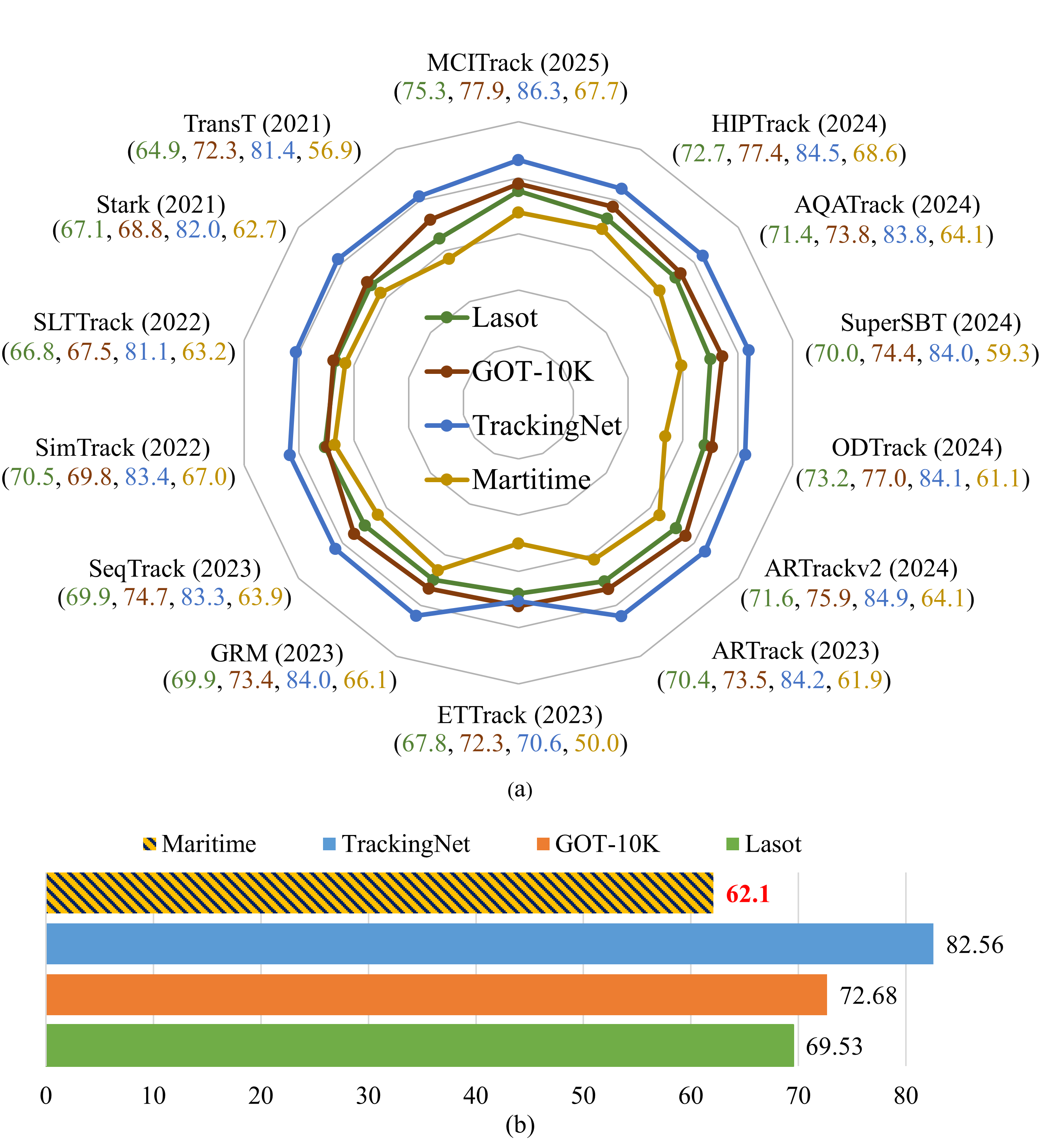} 
  \caption{Visual comparison of tracking algorithm performance: (a) demonstrates individual tracker performance variations across four datasets through AUC scores, while (b) presents aggregated performance by showing average AUC scores for each dataset category.}
  \label{fig:Trackers_Performance}
\end{figure}

In recent years, several representative paradigms have significantly advanced VOT performance \cite{siameseFC,TransT,VLM_Tracking}.
These include Discriminative Correlation Filters (DCF), Siamese-network-based trackers \cite{siameseFC, SiamMASK, SiamBAN, SiamRPN++}, Vision Transformer (ViT)-based models \cite{TransT, SeqTrack,TrackAAAI}, and more recently, Vision-Language Models (VLMs) \cite{AITrack,VLM_Tracking}.
These advancements have been greatly facilitated by the availability of large-scale, publicly available benchmark datasets. Notable examples include LaSOT \cite{LaSOT}, the VOT series (spanning 2014-2024 with annual challenges) \cite{VOTLT2019,VOTLT20128,VOTLT2020,VOTLT2021,VOTLT2022,VOT2023}, GOT-10K \cite{Got10K}, and TrackingNet \cite{TrackingNet}, which provide standardized platforms for rigorous evaluation and comparison (Fig.\ref{fig:Trackers_Performance} (a)).

While existing benchmark VOT datasets such as LaSOT \cite{Lasot_Journal}, GOT-10k \cite{Got10K}, and TNL2K \cite{TNL2k} among others, have significantly advanced the generic object tracking, they are not well suited for maritime environments, where target visual appearance differs substantially from terrestrial settings \cite{Survey}.
Consequently, State-Of-The-Art (VOT) visual trackers developed and evaluated solely on these benchmarks may fail to perform reliably in real-world maritime applications like tracking (Fig.\ref{fig:Trackers_Performance} (b)).

The maritime environment presents a unique set of challenges that distinguish it significantly from more commonly addressed terrestrial scenarios \cite{Survey, Waseem7}.
A primary difficulty lies in the \textit{dynamic and cluttered backgrounds}. 
The water surface is in constant motion, exhibiting varying wave patterns and reflections, which create rapidly changing and often highly cluttered target objects, unlike the relatively stable backgrounds of the terrestrial surface \cite{Survey, Waseem5}.
Another key challenge is the variability in \textit{lighting and weather conditions} \cite{SMD}. 
Maritime operations must function under extreme illumination variations. 
These range from intense sunlight with glare and reflections to low-light conditions at dawn, dusk, or night. 
Adverse weather, such as fog, rain, and mist, further degrades visibility \cite{VBAN}.
These conditions vary the visual appearance of the target objects and pose substantial issues for maritime VOT.

In addition to the aforementioned challenges, maritime tracking also suffers from substantial \textit{variability in object appearance}.
Maritime vessels encompass a vast range of types, sizes, and structures, from small, fast-moving boats to large, slow cargo ships. 
Their visual appearance changes dramatically depending on distance, viewing angle, and their orientation, influenced by the sea state. The target objects may experience partial occlusion by waves or other vessels, or even structures like piers, further varying their appearance and making consistent VOT difficult \cite{vessel_occlusion}. 
The \textit{motion patterns of maritime objects} are also often more complex and less predictable than those of vehicles following defined lanes, involving varied speeds and rapid changes in direction \cite{Random_motion}. 
Finally, tracking in maritime settings frequently involves significant \textit{scale variations} as objects may appear very small at long distances and grow substantially in size as they approach, requiring trackers capable of robustly handling these appearance variations.
\textit{These substantial maritime VOT challenges underscore the critical need for dedicated datasets specifically curated to capture the multifaceted challenges of the maritime environment.}
Without such tailored resources, the development and rigorous benchmarking of accurate and reliable maritime object tracking systems remain considerably hindered.

To bridge this gap and facilitate advancements in maritime VOT, in this work, we propose a novel maritime object tracking dataset. 
This maritime domain-specific dataset reflects the diverse and challenging conditions encountered in real-world maritime surveillance and tracking.
Our dataset aims to compile a comprehensive collection of video sequences featuring a variety of maritime vessels operating under a wide spectrum of environmental conditions, including different lighting, weather states, sea states, and levels of background clutter. 
A key component of this dataset is the inclusion of high-quality, frame-by-frame annotations, such as precise bounding boxes, specifically designed to support the training, validation, and rigorous evaluation of object tracking algorithms within maritime scenarios.
By offering a rich, diverse, and representative dataset of maritime scenes and objects, we intend to provide the necessary resource to accelerate research and development toward more robust, accurate, and reliable maritime object tracking, thus enhancing safety, security, and operational efficiency in the maritime domain.

\begin{figure*}[t]
  \centering
  \includegraphics[width=\textwidth]{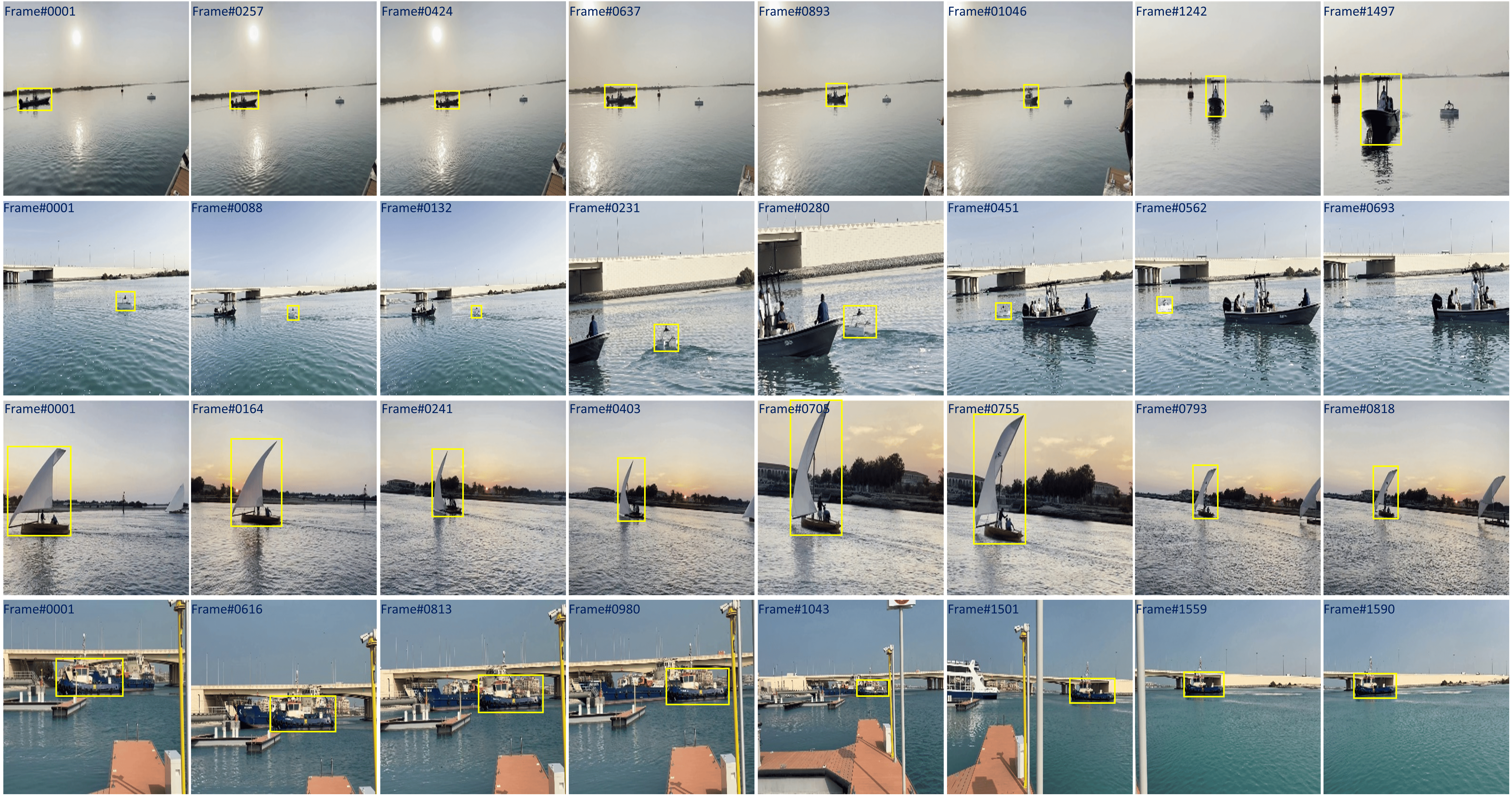} 
  \caption{Sample frames from the maritime object tracking dataset. Each row corresponds to a different object category: 1-boat, 2-USV, 3-sailboat, and 4-ship. For each category, eight frames from different timestamps within the video sequences are shown, illustrating the appearance variation and motion dynamics of the objects across time.}
  \label{fig:Data_Samples}
\end{figure*}

Our maritime VOT dataset contains 182 video sequences with approximately 150K frames.
The dataset contains four distinct maritime objects categories including boat, ship, sailboat, and Unmanned Surface Vehicle (USV) shown in Fig. \ref{fig:Data_Samples}.
It presents core maritime domain-specific challenges, such as \textit{illumination variations} due to water reflections, \textit{target appearance variation} caused by environmental factors, \textit{low contrast objects} blending with the water, and \textit{background clutter} from waves or other marine objects. 
In addition, it also contains generic VOT core challenges challenges including \textit{occlusion}, \textit{scale variation}, \textit{motion blur}, \textit{partial visibility}, and \textit{low resolution}.
These challenges are common in publicly available VOT benchmark datasets.
The dataset contains annotated videos of various maritime tracking scenarios, including coastal surveillance, harbor monitoring, and open-sea vessel tracking.
Our dataset includes a diverse range of vessel types and environmental conditions, providing a comprehensive benchmark for maritime tracking. 
A dedicated maritime tracking dataset is essential, as no existing resources fully capture the inherent challenges of this domain. 
By providing a dataset targeting maritime environment, the tracking community may develop and refine visual trackers specifically designed to meet the distinct demands of maritime surveillance, bridging the gap between general-purpose tracking methods and specialized, real-world applications.
We evaluated the performance of 14 existing SOTA trackers in our proposed maritime dataset using different protocols (Fig.\ref{fig:Trackers_Performance}).
Our analysis reveals that most SOTA trackers suffer from performance degradation while addressing the core challenges of maritime tracking. 
As shown in Fig. \ref{fig:Trackers_Performance} (a), SOTA trackers performance is lower on our dataset compared to their performance on other VOT benchmark datasets.
This is primarily due to the maritime-specific challenges. T
This highlights the pressing need for a dedicated maritime tracking dataset to improve and evaluate trackers under real-world maritime conditions.

Our main contributions in this paper are as follows: 

\begin{enumerate}
   \item \textbf{A Novel Maritime Tracking Dataset:} We introduce the first large-scale maritime VOT dataset comprising 182 annotated video sequences and approximately 150,000 frames. 
   The dataset includes four key vessel categories—boat, ship, sailboat, and Unmanned Surface Vehicle (USV)—captured across diverse operational scenarios such as coastal surveillance, harbor monitoring, and open-sea tracking.
 \item \textbf{Comprehensive Maritime VOT Challenges:} Our dataset is uniquely designed to capture maritime domain-specific challenges, such as illumination variation due to water reflections, target-background low contrast, and environmental clutter. In addition, the generic VOT challenges include occlusion, motion blur, and scale variation. 
 These tracking challenges enable our dataset for robust benchmarking by evaluating trackers in real-world maritime environments.  
\item \textbf{Extensive Benchmarking and Empirical Insights:} We provide an extensive performance evaluation of 14 SOTA visual trackers on the proposed dataset using standard protocols.
Our analysis reveals significant performance degradation under maritime-specific conditions, clearly demonstrating the limitations of existing trackers and highlighting the pressing need for domain-adapted tracking solutions.
 
\end{enumerate}

The rest of this work is organized as follows: Section \ref{sec:r-work} reviews related work, including open-air public datasets, maritime-specific datasets, and recent advancements in VOT algorithms. Section \ref{sec:Proposed_Dataset} provides a detailed discussion of our proposed maritime dataset. Section \ref{sec:exp} presents rigorous benchmarking and experimental evaluations, while Section \ref{sec:Analysis} offers an in-depth analysis of tracker performance across various dataset attributes. Finally, Section \ref{sec:conclusion} concludes the paper and highlights potential directions for future research.

\vspace{-4mm}
\section{Related Work}\label{sec:r-work}

In this section, we provide an overview of the broad range of visual object tracking dataset benchmarks, discuss the scarcity of dedicated maritime datasets, and summarize recent advancements in state-of-the-art tracking algorithms.

\vspace{-4.5mm}
\subsection{VOT Benchmark Datasets}

In the past decade, many open-air VOT benchmark datasets have been proposed in the literature \cite{Dataset_Survey}. These datasets can be broadly categorized into short-term and long-term tracking datasets. Short-term datasets typically consist of a limited number of frames (usually fewer than 600) where the target remains visible throughout the sequence, without requiring re-detection. In contrast, long-term datasets contain sequences that are significantly longer often exceeding 600 frames and are designed for cases where the target may disappear and later reappear, requiring the tracker to incorporate re-detection mechanisms \cite{Lasot_Journal}.

Popular short-term tracking datasets include OTB13 \cite{OTB13}, OTB15 \cite{OTB15}, TC128 \cite{TC128}, the VOT challenge series (latest version available in 2023) \cite{VOT2023}, NUS-PRO \cite{NUS-PRO}, GOT-10K \cite{Got10K},  and TrackingNet \cite{TrackingNet}. These datasets primarily evaluate a tracker’s ability to handle occlusion, motion blur, illumination variations, and scale changes. They provide extensive benchmark comparisons and are widely used in the tracking community.

On the other hand, long-term tracking datasets focus on more complex scenarios where the target may leave the field of view or undergo full occlusion. Notable long-term datasets include UAV123 \cite{UAV123}, UAV20L \cite{UAV123}, LaSOT \cite{Lasot_Journal}, NFS \cite{NFS}, CDTB \cite{CTDB}, and the VOT-LT challenge series from 2018 to 2022 \cite{VOTLT20128, VOTLT2019, VOTLT2021, VOTLT2022}. These datasets introduce additional challenges such as target re-detection, prolonged occlusions, and extreme motion variations.

The availability of these datasets has significantly contributed to the advancement of SOTA tracking performance. Many datasets provide online evaluation platforms for standardized comparisons, such as those offered by TrackingNet, GOT-10K, and the VOT series. Additionally, attribute annotations such as occlusion, motion blur, scale variation, and illumination changes allow researchers to analyze tracker performance under specific challenges. Performance leaderboards for datasets like TrackingNet, GOT-10K, and VOT further facilitate continuous improvements in tracking algorithms. Annual challenges such as the VOT competition, organized since 2014, continue to motivate researchers to develop more robust tracking models.
\vspace{-4.5mm}
\subsection{Maritime Datasets}
Despite advancements in open-air tracking, there is a significant lack of dedicated benchmark datasets for maritime object tracking. While several object detection datasets are available in the maritime domain, such as the SeaDronesSee dataset for detecting humans in open water \cite{SEADRONESEE}, POLARIS for multi-modal maritime data consisting of just 6 sequences \cite{POLARIS}, and the Singapore Maritime Dataset (SMD), which includes limited tracking data with 36 video sequences \cite{SMD}, none fully address the inherent challenges of maritime tracking. Most of the comprehensive existing datasets focus on terrestrial and aerial environments, highlighting the need for domain-specific datasets for maritime challenges.

\vspace{-4.5mm}

\subsection{SOTA Visual Trackers}
The mainstream VOT algorithms are developed based on RGB videos and have been significantly boosted by deep learning techniques in recent years. Convolutional neural networks (CNNs) were initially adopted for feature extraction and learning. Specifically, the MDNet series \cite{MDNet} extracts deep features using three convolutional layers and learns domain-specific layers for tracking. Xu et al. \cite{Xu2023} proposed a spatial-time discrimination model based on affine subspace for VOT. The SiamFC \cite{siameseFC} and SINT \cite{SINT} were among the first to utilize Siamese fully convolutional neural networks and Siamese instance matching for tracking, respectively. In addition, a topology-aware universal adversarial attack method against 3D object tracking was proposed by \cite{topology}. Gradually, Siamese network-based trackers have become mainstream, with many representative trackers being introduced, such as SiamRPN++ \cite{SiamRPN++}, SiamMask \cite{SiamMASK}, SiamBAN \cite{SiamBAN}, Ocean \cite{OCEAN}, LTM \cite{LTM}, ATOM \cite{ATOM}, DiMP \cite{DIMP}, and PrDiMP \cite{PRDIMP}.

Inspired by the success of self-attention mechanisms and Transformer networks in natural language processing, researchers have also explored Transformers for VOT \cite{TRDiMP,TrT,TOMP,AIATRACK,SimTrack,OSTRACK}. For example, Wang et al. \cite{TRDiMP} proposed TrDiMP, which integrates transformers with tracking tasks, exploiting temporal context for robust visual tracking. Chen et al. \cite{TrT} introduced TransT, a novel attention-based feature fusion network combined with a Siamese structured tracking approach. ToMP \cite{TOMP} proposed a Transformer-based model prediction module, enhancing target prediction capabilities due to Transformer's inductive bias in capturing global relationships. Gao et al. \cite{AIATRACK} proposed AiATrack, introducing a universal feature extraction and information propagation module based on Transformers. SimTrack \cite{SimTrack}, proposed by Chen et al., utilizes Transformers as the backbone for joint feature extraction and interaction. Ye et al. \cite{OSTRACK} proposed OSTrack, designing a one-stream tracking framework to replace the complex dual-stream architecture. 

In another approach, knowledge distillation has been widely studied for learning student networks to achieve efficient and accurate inference. Deng et al. \cite{DENG} provide explicit feature-level supervision for event stream learning by distilling knowledge from the image domain. For tracking tasks, Shen et al. \cite{SHEN} propose distilling large Siamese trackers using a teacher-student knowledge distillation model to develop small, fast, and accurate trackers. Chen et al. \cite{CHEN} focus on learning a lightweight student correlation filter-based tracker by distilling a pre-trained deep convolutional neural network. Zhuang et al. \cite{Zhuang} introduce ensemble learning (EL) into the Siamese tracking framework, treating two Siamese networks as students to enable collaborative learning. Sun et al. \cite{SUN} conduct cross-modal distillation for TIR tracking from RGB modality on unlabeled paired RGB-TIR data. Wang et al. \cite{WANG} distill the CNN model pre-trained from image classification datasets into a lightweight student network for fast correlation filter trackers. Zhao et al. \cite{ZHAO} propose a distillation-ensemble-selection framework to address the conflict between tracking efficiency and model complexity. Ge et al.  \cite{GE} propose channel distillation for correlation filter trackers, accurately mining better channels and alleviating the influence of noisy channels. 

\vspace{-5mm}
\section{Proposed Maritime Dataset}\label{sec:Proposed_Dataset}
This research work aims to introduce a domain-specific benchmark for the training and evaluation of object-tracking methods in maritime environments. Our proposed dataset is large-scale, featuring high-quality annotations. The sequences presents a long-term and short-term tracking challenges, with diverse target objects representing a wide range of maritime scenarios.
\vspace{-5mm}
\subsection{Dataset Details:}
As shown in Table \ref{tab:Dataset_Summary}, 
our proposed maritime VOT dataset consists of 182 sequences of moving objects with an average length of ~863 frames per video, captured at 30 and 60 fps. The sequences vary in frame resolution, with the minimum and maximum resolutions being 1024 $\times$ 1024 and 1920 $\times$ 1440, respectively. In total, the dataset includes 150,058 annotated frames, with sequence lengths ranging from a minimum of 82 frames to a maximum of 4747 frames. Among these, 57 sequences have fewer than 300 frames, presenting short-term tracking challenges, while 125 sequences feature longer durations, offering long-term tracking challenges. The dataset includes 9 distinct tracking attributes, covering both maritime-specific characteristics as well as general attributes commonly found in other tracking datasets. It features 4 diverse target boat categories, with varying sizes and motion patterns as shown in Fig. \ref{fig:Data_Samples} .

\begin{table}[ht]
\centering
\caption{Summary of the Proposed Maritime VOT Dataset}
\begin{tabular}{ l l }
\hline
\textbf{Characteristic} & \textbf{Value} \\
\hline
Total sequences & 182 \\
Frame rates & 30 fps \& 60 fps \\
Total annotated frames & 150{,}058 \\
Minimum Resolution & 1024×1024 \\
Maximum Resolution & 1920×1440 \\
Minimum number of frames & 82  \\
Maximum number of frames & 4747  \\
Average number of frames & 824 \\
Short-term sequences & 57 \\
Long-term sequences & 125 \\
Tracking attributes & 9  \\
Object categories & 4 \\
\hline
\end{tabular}
\label{tab:Dataset_Summary}
\end{table}
\vspace{-4.5mm}
\subsection{Dataset Collection}
We have compiled a comprehensive dataset by recording extensive video footage of vessels and ships navigating through maritime environments. The recordings were systematically captured using two distinct camera setups: an onshore static camera and an offshore dynamic camera mounted on an Unmanned Surface Vessel (USV) \cite{US}. This dual-camera approach significantly enhances dataset diversity, providing robust data that captures a wide range of realistic maritime conditions.

The onshore camera contributes critical challenges, such as significant scale variations due to varying distances between the camera and vessels, frequent occlusions caused by environmental factors such as passing ships or natural obstructions, and perspective distortions arising from viewing angles. In addition, this setup faces challenges like varying weather conditions that impact visibility and image quality. 
In contrast, the dynamic camera installed on the USV captures a distinct set of challenges. These include significant illumination variations caused by reflections from sunlight on the water surface, glare, and rapid changes in lighting conditions throughout the day. Moreover, motion blur caused by the dynamic movement of the vessel, water splashes hitting the camera lens, vibrations from waves, and continuous positional shifts create additional complexity. Furthermore, the onboard camera faces challenges related to stabilization difficulties during rough sea conditions and inconsistent frame composition as a result of the USV's unpredictable motion.
The combination of onshore and offshore USV-mounted cameras ensures the dataset encompasses a broad spectrum of maritime operational scenarios. That will require highly robust tracking algorithms to address the challenges of real-world maritime vessel monitoring and tracking applications.

\begin{figure*}[t]
  \centering
  \includegraphics[width=\textwidth]{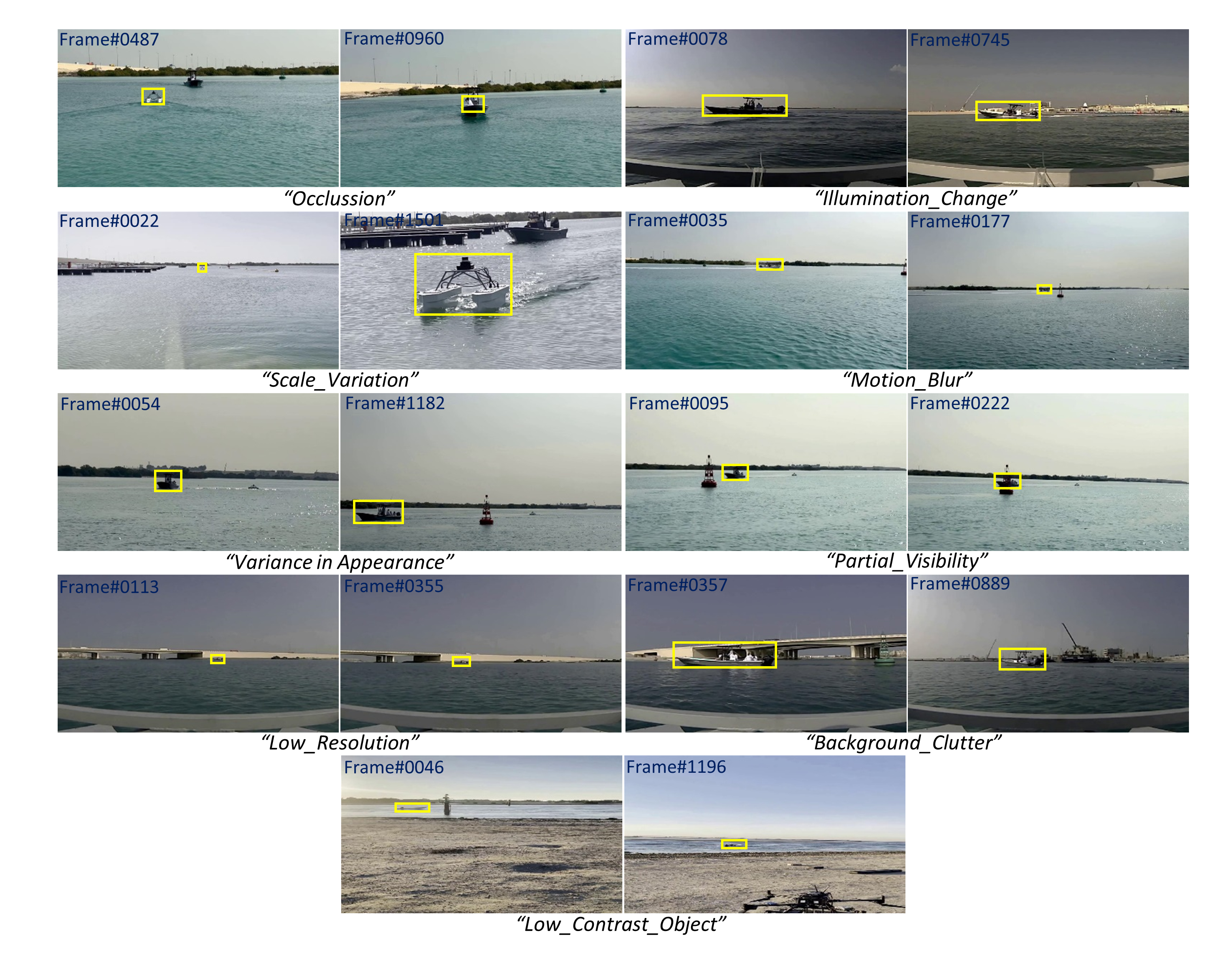} 
  \caption{Visual examples of nine challenging attributes in MVTD: Occlusion, Illumination Change, Scale Variation, Motion Blur, Variance in Appearance, Partial Visibility, Low Resolution, Background Clutter, and Low Contrast Object. These attributes highlight diverse visual difficulties affecting tracking performance.}

  \label{fig:Att}
\end{figure*}

\begin{table*}[htbp]
\centering
\caption{Overview of recent visual object tracking methods, highlighting their backbone architectures and key contributions.}
\label{tab:Trackers_Summary}
\begin{tabular}{l@{\hspace{1.5em}}r l p{11cm}}
\hline
\multicolumn{2}{l}{\textbf{Trackers}} & \textbf{Backbone} & \textbf{Description} \\
\hline
MCITrack & (2025)\cite{TrackAAAI} & Fast-iTPN & Introduces a hidden state-based contextual fusion module leveraging Mamba layers to store and transmit rich video-level context, significantly improving tracking accuracy without extra tokens or high computational cost. \\ 
HIPTrack & (2024)\cite{cai2024} & ViT-B & Introduces a historical prompt network that encodes refined foreground masks and past visual features, enabling accurate, efficient tracking without full model retraining. \\ 
AQATrack & (2024)\cite{AQA} & HiViT-B & Uses learnable autoregressive queries and a spatio-temporal fusion module to adaptively model appearance changes, avoiding handcrafted update rules while enhancing temporal reasoning and robustness. \\
SuperSBT & (2024)\cite{SuperSBT} & Custom & Proposes a single-branch transformer pipeline with joint feature extraction and cross-image interaction, enhanced by hierarchical design, unified relation modeling, and masked image pretraining for faster and more accurate tracking. \\ 
ODTrack & (2024)\cite{OD} & ViT-B & Proposed an online token propagation to model video-level associations, compressing target features into iterative prompt tokens that guide frame-to-frame predictions through dense spatiotemporal reasoning. \\
ARTrackv2 & (2024)\cite{ARv2} & ViT-B & Jointly models trajectory prediction and appearance evolution through inter-frame autoregression, using spatio-temporal prompts to propagate target dynamics and features across frames. \\ 
ARTrack & (2023)\cite{AR} & ViT-B & Autoregressive tracker that predicts bounding box coordinates sequentially using previous states as spatio-temporal prompts in encoder-decoder framework. \\ 
ETTrack & (2023)\cite{ETTrack} & LT-Mobile & Introduces Exemplar Transformer with global query and learned exemplar keys to replace convolutional heads in Siamese tracking framework. \\ 
GRM & (2023)\cite{GRM} & ViT-B & Introduces a generalized relation modeling method with adaptive token division, unifying one-stream and two-stream tracking pipelines. \\ 
SeqTrack & (2023)\cite{SeqTrack} & ViT-B & Introduces a sequence-to-sequence framework that models tracking as a generation task, utilizing an encoder-decoder transformer for autoregressive bounding box prediction without complex head networks. \\ 
SimTrack & (2022)\cite{SimTrack} & Custom & A one-branch transformer backbone for joint feature learning and interaction with introduction of a foveal window strategy to reduce information loss from down-sampling and enhance localization accuracy. \\ 
SLTTrack & (2022)\cite{SLTtrack} & ResNet-50 & Sequence-level training strategy utilizing reinforcement learning to align data distribution and objectives with real-world tracking scenarios. \\ 
STARK & (2021)\cite{Stark} & ResNet-50 & Encoder-decoder Transformer architecture modeling spatio-temporal dependencies for direct bounding box prediction without post-processing. \\ 
TransT & (2021)\cite{TransT} & ResNet-50 & Attention-based feature fusion using self- and cross-attention modules to replace traditional correlation operations for better semantic integration and robustness. \\ 
\hline
\end{tabular}
\end{table*}

\vspace{-5mm}
\subsection{Dataset Annotation}
For annotation, we follow a simple yet consistent annotation protocol. In the first frame of each sequence, a precise bounding box is manually drawn to fit the target object. In subsequent frames, a rough bounding box is generated using the Computer Vision Annotation Tool (CVAT). This is then manually checked and adjusted by a team of M.Sc. and Ph.D. students having expertise in image processing domain to ensure it fits the target object as closely as possible while excluding as much background as possible. If the target is partially occluded or out of view, the frame is marked with an “out-of-view” or “full occlusion” label.

After the bounding boxes are annotated, they are reviewed by a validation team to check for accuracy. If any annotation does not fully cover the target object or includes too much background, it is sent back for revision. For certain challenging objects, like sailboat with long and deformable parts, specific rules are followed to exclude irrelevant areas.

After the team makes corrections, the final annotation for each frame is computed by averaging the corrected bounding boxes. This process ensures consistent and precise annotations across all video sequences in the dataset.

\subsection{Attributes}
In order to evaluate the performance of tracking algorithms under various challenges, each video sequence in our maritime dataset is categorized under 9 distinct attributes. These attributes capture specific characteristics of the video that may pose particular challenges to the tracker. As shown in Fig \ref{fig:Att}, the following attributes are considered for the evaluation:

\textbf{1-Occlussion:} This occurs when the target object is temporarily blocked by another object or obstruction, making it difficult for the tracker to maintain the correct identification.

\textbf{2-Illumination Changes:} Refers to variations in lighting conditions, such as light reflections from the water surface. This can change the appearance of the target, making tracking difficult under changing lighting conditions.

\textbf{3-Scale Variation:} When the target moves closer to or farther away from the camera, its size changes, posing a challenge for trackers. The tracker may struggle to adjust to these changes and either miss the target or misplace the bounding box.

\textbf{4-Motion Blur:} Rapid movement of either the target or camera leads to a loss of image sharpness, making it hard for the tracker to detect and follow the target accurately. This is common in fast-moving objects or turbulent water conditions.

\textbf{5-Variation in Appearance:} This occurs when the target undergoes significant changes in appearance due to environmental factors, such as water reflection or target deformations. These changes make it difficult for trackers that rely on appearance models to maintain accurate tracking.

\textbf{6-Partial Visibility:} The target may not always be fully visible within the frame due to occlusions from other objects or environmental factors. When the target is partially obstructed or only visible from certain angles, the tracker must predict its position to avoid losing track.

\textbf{7-Low-Resolution:} Low-quality videos with lower resolution make it difficult for the tracker to detect fine details, leading to decreased tracking accuracy. This is especially common in large maritime environments, where the target may appear as a small or blurry object.

\textbf{8-Background Clutter:} Complex or visually dense backgrounds, such as the open sea or coastal areas, make it hard for the tracker to isolate the target from the surrounding environment. Other moving vessels or marine objects may also distract or confuse the tracker, reducing its effectiveness.

\textbf{9-Low Contrast Objects:} Targets that have a color or texture that closely matches the water or surrounding environment are difficult for trackers to identify, especially in low light or murky water. Low contrast objects blend in with the background, requiring specialized techniques to enhance visibility for tracking.
\vspace{-5mm}
\subsection{Trackers}
To thoroughly assess the robustness and generalization capabilities of existing tracking models in maritime environments, we evaluate a comprehensive set of 14 SOTA trackers on our proposed dataset. These trackers are widely used in the visual tracking community and have demonstrated strong performance across various benchmarks. Each tracker is briefly described in table \ref{tab:Trackers_Summary}, outlining its core methodology and relevance to our evaluation.

\begin{figure*}[p]
  \centering
  \includegraphics[width=\textwidth, height=20cm]{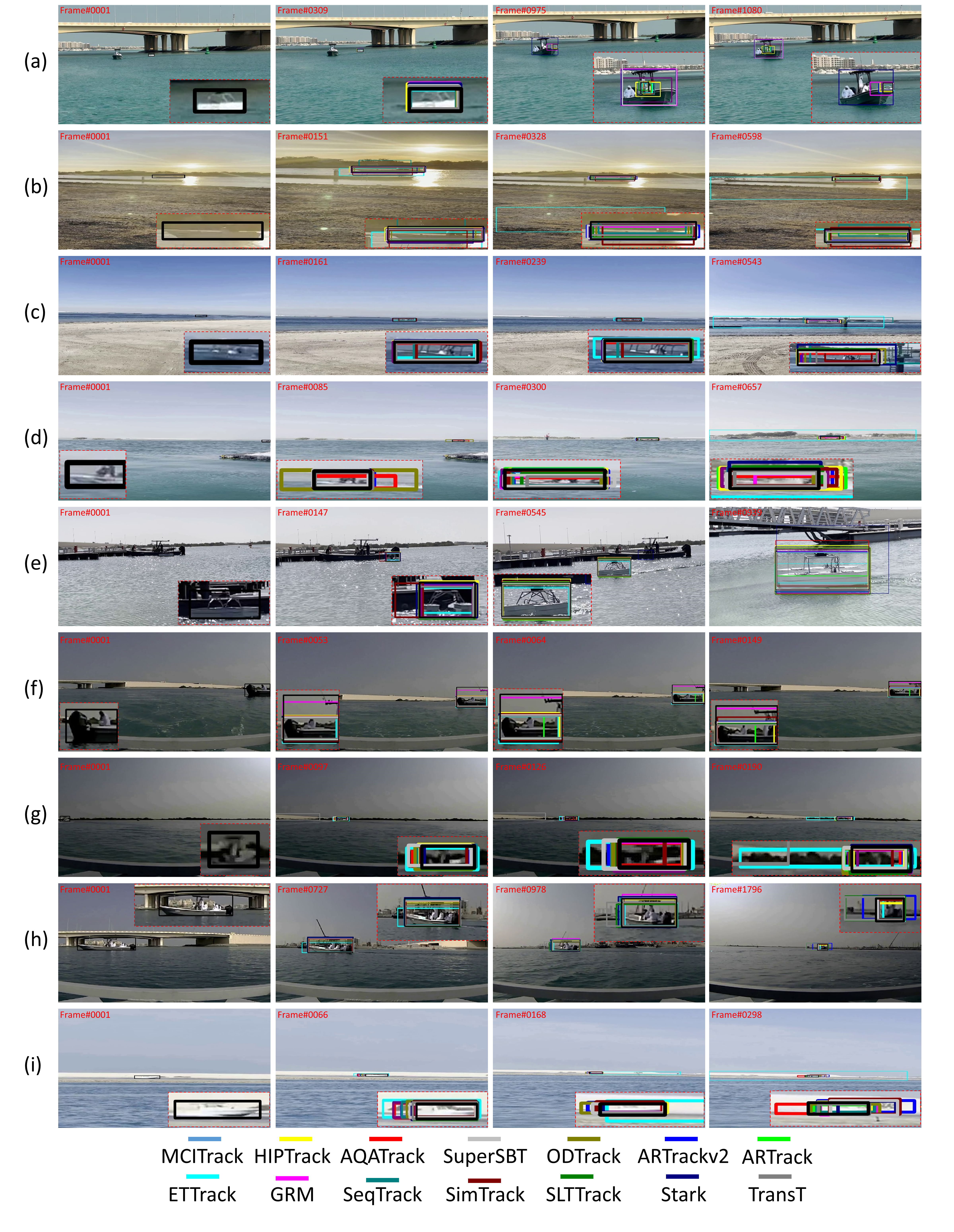}
  \vspace{-5mm}
  \caption{ Sample images from the maritime dataset showcasing tracked target objects across diverse attributes and tracking challenges. The figure contains 9 rows (a$–$i) corresponding to different attributes: Occlusion, Illumination Change, Scale Variation, Motion Blur, Variance in Appearance, Partial Visibility, Low Resolution, Background Clutter, and Low-Contrast Objects. Each image displays bounding boxes predicted by 14 state-of-the-art (SOTA) trackers, color-coded as indicated in the legend whereas black color is the groundtruth color. A zoomed part of tracked object is also added for better visibility}
  \label{fig:Attributes}
\end{figure*}

\vspace{-2mm}
\section{Experimentation}\label{sec:exp}

\subsection{Evaluation Protocols}
We evaluated the proposed maritime dataset using two distinct evaluation protocols to provide a detailed analysis of the results:

\begin{itemize}
    \item \textbf{Pretrained Model Evaluation (Protocol I):} This protocol evaluates the generalization capability of different trackers, trained on existing open-air datasets, to the maritime dataset. Pretrained trackers are tested on the complete test set of the proposed maritime dataset. For this protocol, we report the results for the entire dataset and, to facilitate comparison, also provide the results for the testing split of the dataset.
    
    \item \textbf{Fine-Tuning Evaluation (Protocol II):} In this protocol, trackers are fine-tuned using the training split of the maritime dataset and evaluated on the predefined testing split. The dataset is divided into training and testing splits, ensuring that all object categories are present in both splits with no overlap between them.
\end{itemize}
\vspace{-5mm}
\subsection{Evaluation Metrics}
We evaluate the performance of the trackers using the following metrics:

\begin{itemize}
    \item \textbf{Precision:} Precision measures the accuracy of the tracker by calculating the Euclidean distance between the center of the ground truth bounding box and the center of the predicted bounding box. A frame is considered correct if the Euclidean distance is less than a predefined threshold. The precision is then plotted by varying the threshold from 0 to 50 pixels, and the average precision is reported.
    
    \item \textbf{Normalized Precision:} To address scale variations of the target, we also compute the normalized precision, which takes into account the diagonal of the target object's bounding box. This metric provides a more accurate measure of precision, considering object size variations.
    
    \item \textbf{Success Rate:} The success rate is computed as the ratio of the number of successfully tracked frames (i.e., frames where the intersection-over-union (IoU) between the ground truth bounding box and the tracking result exceeds a predefined threshold, typically 0.5) to the total number of frames in a sequence.
\end{itemize}
\vspace{-5mm}
\subsection{Implementation Details}
For Protocol I, we use pre-trained trackers from open-air datasets on the maritime dataset. The default parameters, as recommended by the original authors, are used for each tracker.

For Protocol II, top 5 performing trackers are selected and fine-tuned using the raw training split of the maritime dataset. Inference is performed on the raw testing split of the maritime dataset.

All experiments are conducted on a workstation equipped with an NVIDIA RTX A6000 GPU with 48GB of memory, running CUDA 12.2, and using PyTorch 1.11.0+cu102 on Ubuntu 20.04. The system is powered by Intel Xeon Gold 6230 CPU with 40 physical cores and 80 threads. It has a total of 256 GB of memory. The trackers are implemented using the official source codes provided by the respective authors.

\begin{figure*}
\begin{minipage}{0.33\textwidth}
  \centering
  \includegraphics[width=\linewidth]{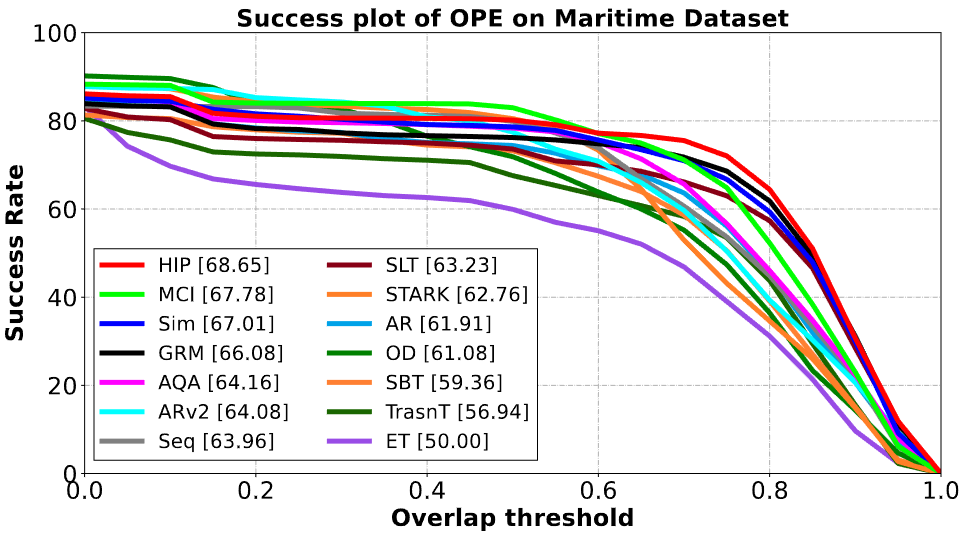}
  \subcaption{(a)}
  \label{fig:first}
\end{minipage}%
\hfill 
\begin{minipage}{0.33\textwidth}
  \centering
  \includegraphics[width=\linewidth]{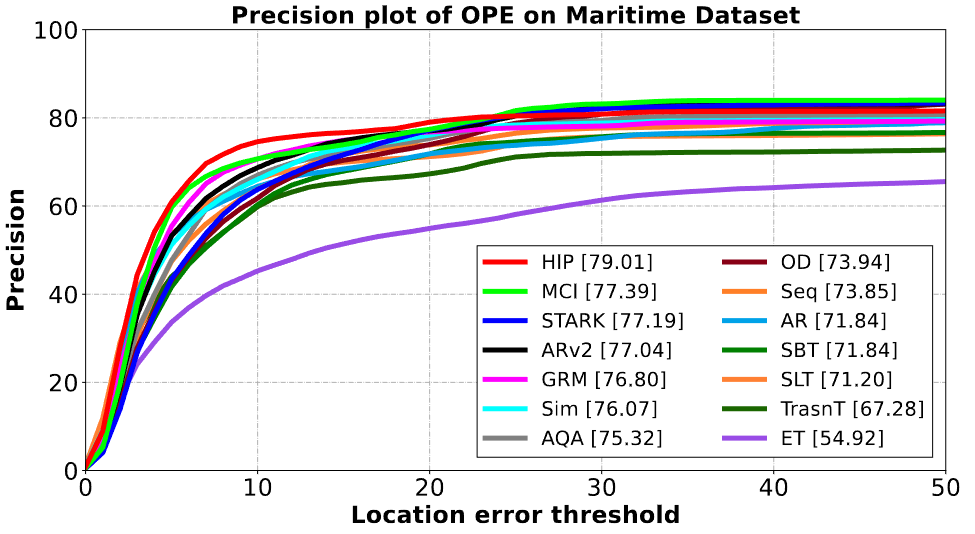}
  \subcaption{(b)}
  \label{fig:second}
\end{minipage}%
\hfill
\begin{minipage}{0.33\textwidth}
  \centering
  \includegraphics[width=\linewidth]{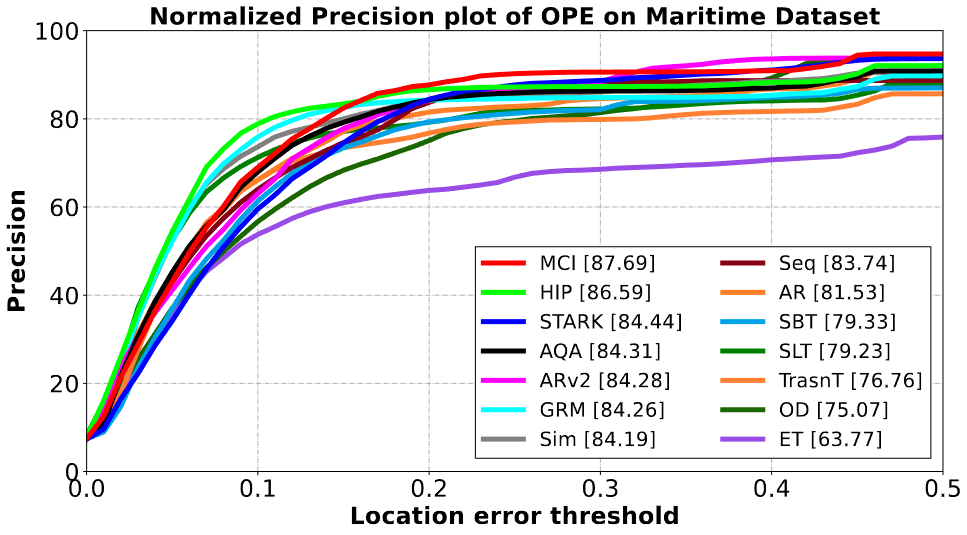}
  \subcaption{(c)}
  \label{fig:third}
\end{minipage}%
  \caption{Performance evaluation on Protocol I showing: (a) Success plot with AUC scores measuring bounding box overlap, (b) Precision plot showing center location error in pixels, and (c) Normalized precision plot accounting for target size variations. Higher values indicate better performance in all metrics.}
  \label{fig:Pretrained_results}
\end{figure*}

\vspace{-2mm}
\section{Analysis of Results}\label{sec:Analysis}

\subsection{Protocol I Analysis}

The overall performance in terms of  success rate, precision and normalized precision plots of the evaluated trackers is shown in Fig \ref{fig:Pretrained_results} (a), (b), and (c). The evaluation reveals distinct differences in the tracking capabilities of various algorithms when evaluated on the maritime dataset. HIP came out as the top-performing tracker, achieving the highest success rate of 68.65\%, which suggests its robustness in maintaining target localization across sequences. MCI (67.78\%) and GRM (66.08\%) follow closely, demonstrating strong performance in accurately predicting object positions over time. These trackers consistently outperform others, likely due to their ability to handle appearance variations and occlusions effectively.

As shown in table \ref{tab:Pretrained_results} When considering overlap precision at a moderate threshold of 50\% (OP50), MCI outperforms all trackers with an 82.97\% score, reflecting its ability to maintain high IoU with the ground truth. HIP (80.17\%) and STARK (80.49\%) also show competitive results, indicating their reliability in tracking stability. However, as the threshold becomes more high at OP75, performance differences become more evident. HIP (72.02\%) remains the top tracker in this category, followed by GRM (68.58\%) and Sim (66.73\%), suggesting that these trackers can retain accurate localization even with stricter overlap constraints. In contrast, some trackers, such as STARK (43.13\%) and SBT (50.55\%), exhibit a sharp decline, indicating their sensitivity to precise localization requirements.  

In terms of precision, which measures how close the predicted center of the object is to the ground truth, HIP once again leads with 79.01\%, followed by MCI (77.39\%) and GRM (76.80\%). The high precision scores of these trackers suggest that they are particularly effective in minimizing localization errors. Interestingly, despite achieving moderate success rates, Seq (73.85\%), Sim (76.07\%), and AQA (75.32\%) maintain relatively strong precision scores, implying that while they may struggle with long-term robustness, they still perform well in frame-wise localization.

The normalized precision metric provides further insights into how well trackers adapt to variations in object scale. Here, MCI achieves the best performance at 87.69\%, slightly ahead of HIP (86.59\%) and AQA (84.31\%). The high values in this category indicate that these trackers effectively handle size changes, making them more suitable for real-world scenarios where objects undergo scale variations due to perspective changes or movement toward or away from the camera.  

On the other end of the spectrum, some trackers exhibit significantly lower performance across all metrics. ET records the lowest success rate at 50.00\%, with TransT (56.94\%) and SBT (59.36\%) also underperforming. These trackers struggle with maintaining long-term object tracking, likely due to limitations in feature extraction, robustness to occlusions, or adaptability to varying environmental conditions. The ET tracker particularly shows poor OP75 performance (39.02\%), indicating a high tendency for inaccurate localization when stricter thresholds are applied.  

Overall, HIP, MCI, and GRM stand out as the most reliable trackers, consistently achieving high success rates, precision, and overlap precision at different thresholds. Their ability to handle challenging scenarios such as occlusions, scale variations, and fast-moving objects makes them strong candidates for real-world tracking applications. In contrast, lower-performing trackers like ET, TansT, and SBT highlight the challenges faced by models that struggle with robustness and adaptability. This evaluation underscores the importance of designing tracking algorithms that balance success rate, precision, and robustness to environmental variations, ensuring stable and accurate tracking in diverse conditions.

\begin{table}[t]
    \centering
    \caption{Performance comparison on Maritime dataset.}
    \begin{tabular}{lccccc}
        \hline
        \textbf{Tracker} & \textbf{AUC} & \textbf{OP50} & \textbf{OP75} & \textbf{Precision} & \textbf{Norm Precision} \\
        \hline
        AQA \cite{AQA}    & 64.16  & 78.36  & 56.64  & 75.32  & 84.31  \\
        AR \cite{AR}     & 61.91  & 74.41  & 56.09  & 71.84  & 81.53  \\
        ARv2 \cite{ARv2}   & 64.08  & 77.43  & 50.42  & 77.04  & 84.28  \\
        ET \cite{ETTrack}     & 50.00  & 59.92  & 39.02  & 54.92  & 63.77  \\
        GRM \cite{GRM}     & 66.08  & 76.21  & 68.58  & 76.80  & 84.26  \\
        HIP \cite{cai2024}    & 68.65  & 80.17  & 72.02  & 79.01  & 86.59  \\
        MCI \cite{TrackAAAI}     & 67.78  & 82.97  & 64.83  & 77.39  & 87.69  \\
        Seq \cite{SeqTrack}    & 63.96  & 79.88  & 53.64  & 73.85  & 83.74  \\
        Sim \cite{SimTrack}    & 67.01  & 78.63  & 66.73  & 76.07  & 84.19  \\
        SLT \cite{SLTtrack}    & 63.23  & 73.60  & 62.98  & 71.20  & 79.23  \\
        STARK \cite{Stark}  & 62.76  & 80.49  & 43.13  & 77.19  & 84.44  \\
        SBT \cite{SuperSBT}    & 59.36  & 73.20  & 50.55  & 71.84  & 79.33  \\
        OD \cite{OD}  & 61.08  & 71.87  & 47.63  & 73.94  & 75.07  \\
        TransT \cite{TransT} & 56.94  & 67.57  & 53.49  & 67.28  & 76.76  \\
        \hline
    
    \end{tabular}
    \label{tab:Pretrained_results}
\end{table}
\vspace{-5mm}
\subsection{Protocol II Analysis}

\begin{figure*}
\begin{minipage}{0.33\textwidth}
  \centering
  \includegraphics[width=\linewidth]{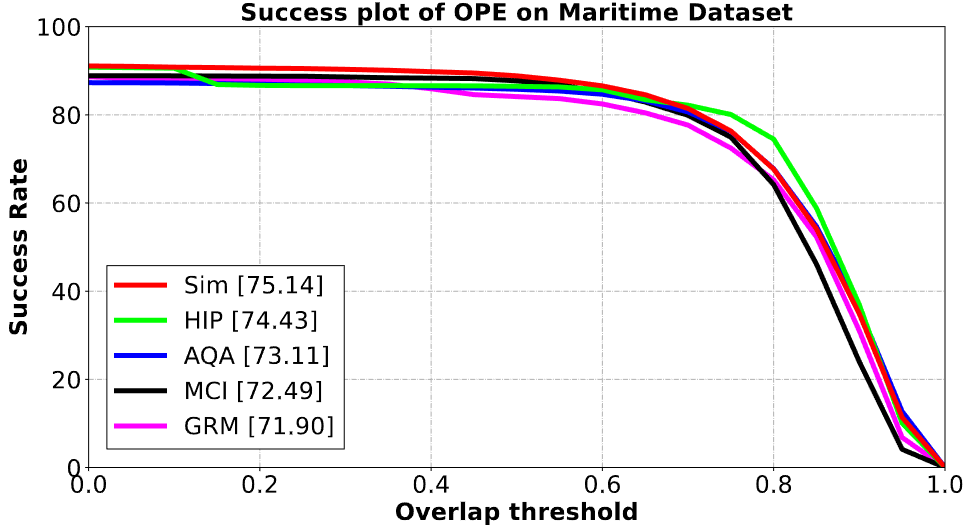}
  \subcaption{(a)}
  \label{fig:first}
\end{minipage}%
\hfill 
\begin{minipage}{0.33\textwidth}
  \centering
  \includegraphics[width=\linewidth]{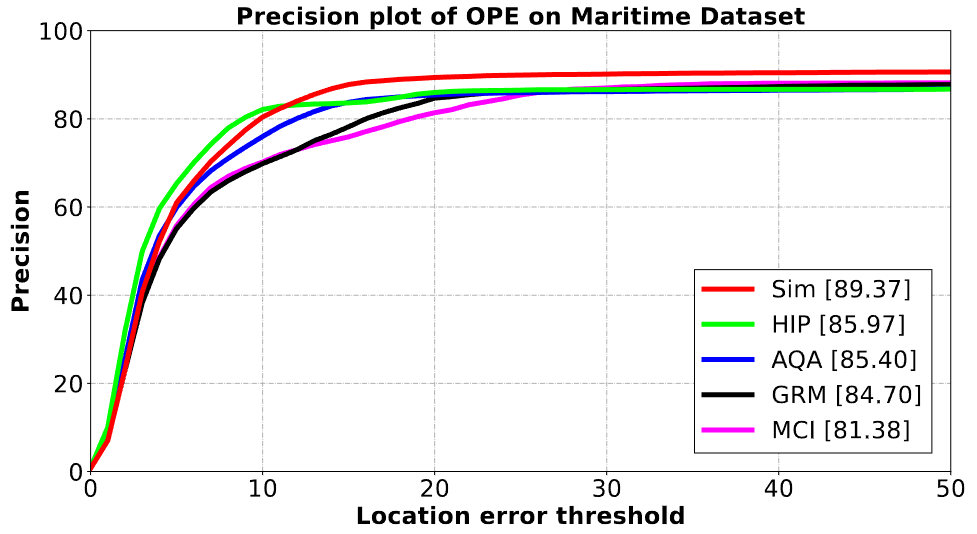}
  \subcaption{(b)}
  \label{fig:second}
\end{minipage}%
\hfill
\begin{minipage}{0.33\textwidth}
  \centering
  \includegraphics[width=\linewidth]{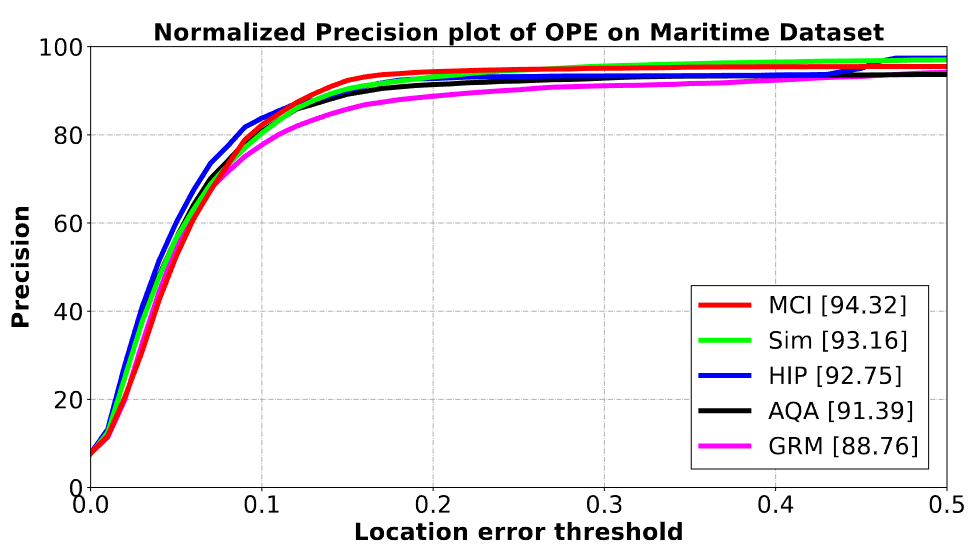}
  \subcaption{(c)}
  \label{fig:third}
\end{minipage}
  \caption{Performance evaluation on Protocol II showing: (a) Success plot with AUC scores measuring bounding box overlap, (b) Precision plot showing center location error in pixels, and (c) Normalized precision plot accounting for target size variations. Higher values indicate better performance in all metrics.}
  \label{fig:FineTuned_Results}
\end{figure*}

Out of the 14 evaluated trackers in protocol I, we selected the five top performing models AQA, GRM, HIP, MCI, and Sim based on their success rate score on protocol I. These models demonstrated strong performance in the initial evaluation and were further fine-tuned by training them on the maritime dataset’s training set before being tested on the test sequences. This fine-tuning process allowed the models to adapt specifically to the inherent challenges of the maritime environment as shown in Fig \ref{fig:FineTuned_Results} (a), (b), and (c).

The results as shown in Table \ref{tab:fine_tuned_results} depicts a substantial improvement across all metrics. This highlights the effectiveness of fine-tuning in enhancing the trackers’ ability to maintain robust and accurate object localization in maritime scenarios.

In terms of the Area Under the Curve (AUC), all fine-tuned models showed noticeable gains. Sim achieved the highest AUC of 75.14\%, followed by HIP (74.43\%) and AQA (73.11\%). These results indicate that fine-tuning on the maritime dataset has helped these trackers better generalize to real-world maritime environment conditions, improving their ability to track objects despite dynamic water backgrounds and changing lighting.

For overlap precision at the moderate OP50 threshold, all fine-tuned models demonstrated significant improvements. Sim achieved the highest OP50 score of 88.84\%, indicating superior tracking stability. MCI and HIP followed with scores of 87.71\% and 86.47\%, respectively. These results confirm that training on maritime-specific data has made these trackers more reliable in maintaining object visibility across sequences.

At OP75 threshold, which requires higher localization accuracy, HIP performed best with an OP75 score of 80.07\%, suggesting that it maintained more precise tracking even in challenging conditions such as waves and partial occlusions. Sim (76.34\%) and AQA (76.25\%) also showed considerable improvements, highlighting their enhanced ability to keep objects tightly localized within their bounding boxes. GRM and MCI followed with competitive OP75 scores of 72.46\% and 74.92\%, respectively.

In terms of precision, Sim achieved the highest precision score of 89.37\%, followed by HIP (85.97\%) and AQA (85.40\%), indicating their increased accuracy in object localization. The normalized precision metric, which accounts for variations in object scale, saw the highest gain in MCI, which scored 94.32\%. This suggests that MCI is particularly effective at handling changes in object size, an essential feature for maritime tracking where objects often appear at different scales due to camera motion and perspective shifts.

Overall, the fine-tuned models exhibited substantial performance gains across all evaluation metrics, demonstrating the benefits of training on domain-specific data. By learning the unique challenges of the maritime environment, such as water reflections, occlusions, and dynamic backgrounds, these models achieved greater robustness and precision in tracking marine objects. The improvements confirm that fine-tuning on a dedicated maritime dataset is a crucial step in enhancing object tracking performance for real-world deployment in marine applications.

\begin{table}[t]
    \centering
    \caption{Performance of fine-tuned top-5 trackers on the MVTD}
    \begin{tabular}{l c c c c c}
        \hline
        \textbf{Tracker} & \textbf{AUC} & \textbf{OP50} & \textbf{OP75} & \textbf{Precision} & \textbf{Norm Precision} \\ 
        \hline
        AQA  & 73.11 & 85.80 & 76.25 & 85.40 & 91.39 \\
        GRM  & 71.90 & 84.13 & 72.46 & 84.70 & 88.76 \\
        HIP  & 74.43 & 86.47 & 80.07 & 85.97 & 92.75 \\
        MCI  & 72.49 & 87.71 & 74.92 & 81.38 & 94.32 \\
        Sim  & 75.14 & 88.84 & 76.34 & 89.37 & 93.16 \\
        \hline
    \end{tabular}
    \label{tab:fine_tuned_results}
\end{table}

\vspace{-5mm}
\subsection{Time Complexity}
In addition to evaluating tracking performance, we also analyze the computational efficiency of the trackers, as summarized in Table \ref{tab:fps_trackers}. The reported FPS values reflect the raw inference speed of each model, measured on a workstation equipped with an NVIDIA RTX A6000 GPU with 48GB RAM.  SimTrack achieved the highest speed at 67.01 FPS, followed by HIPTrack and SuperSBT. In contrast, ETTrack and ARTrack were the slowest.

\begin{table*}[t]
\centering
\scriptsize
\caption{Comparison of inference speed (fps) of all trackers.}
\resizebox{\textwidth}{!}{%
\begin{tabular}{cccccccccccccccc}
\hline
\multicolumn{1}{c|}{Model} & MCITrack & HIPTrack & AQATrack & SuperSBT & ODTrack & ARtrackV2 & ARTrack & ETTrack & GRM & Seqtrack & SimTrack & SLTTrack & Stark & TransT \\
\multicolumn{1}{c|}{Speed(fps)} & 19.99 & 49.26 & 35.91 & 45.35 & 42.34 & 41.27 & 16.28 & 7.27 & 31.24 & 42.59 & 67.01 & 29.06 & 32.35 & 32.50 \\
\hline
\end{tabular}
}
\label{tab:fps_trackers}
\end{table*}

\vspace{-5mm}
\section{Discussion}

The results from our comprehensive evaluation of 14 state-of-the-art visual object trackers on our presented MVTD offer various critical insights into the current limitations of general-purpose tracking algorithms and the demands for effective tracking in complex maritime domains.
\vspace{-5mm}
\subsection{Domain-Specific Challenges}

A significant performance degradation across all trackers in Protocol I highlights a major limitation in existing VOT models is their poor generalization to maritime environments. Despite their success in land-based or open-air benchmarks, these models struggle with challenges specific to maritime scenes, such as low-contrast object boundaries against dynamic water textures, extreme scale variations, and persistent occlusions due to waves or nearby vessels. These results emphasize that maritime tracking is not a mere subset of generic object tracking but a distinct and more demanding domain requiring specialized solutions.
\vspace{-5mm}
\subsection{Fine-Tuning Enables Robust Adaptation}

Protocol II evaluations demonstrated that fine-tuning on domain-specific data significantly increases tracking performance. Trackers such as HIP, Sim, and MCI exhibited up to 10 to 15\% improvements in success and precision metrics after adaptation. This highlights the crucial role of transfer learning and domain adaptation. General-purpose tracking models can excel in specialized fields, but they need to be retrained on relevant datasets. We recommend that future benchmark practices integrate domain adaptation evaluations to assess the actual tracker robustness in diverse environments.
\vspace{-5mm}
\subsection{Attribute-Level Insights}

The attribute-specific representations and results reveal consistent weaknesses in state-of-the-art tracking systems when faced with key maritime challenges. One major issue is the limitation in accurately tracking low-contrast objects; current models rely heavily on appearance-based cues and struggle to distinguish targets that visually blend with the water background. Additionally, scenarios involving motion blur and dynamic occlusions pose significant difficulties, particularly for trackers that lack temporal memory or aggregation mechanisms, such as recurrent networks or attention-based temporal modeling. These conditions cause a rapid and transient change in the appearance of the object, leading to frequent tracking failures. Another persistent challenge is scale variation, where only a few trackers, such as MCI and HIP, demonstrate sufficient adaptability. This highlights the importance of integrating scale-invariant modeling techniques or dynamic feature hierarchies to improve robustness against perspective-induced size changes in maritime tracking scenarios.
\vspace{-4.5mm}
\subsection{Recommendations for Tracker Design}

We suggest the following directions to the tracking research community based on our findings. To enhance the performance of visual object trackers in maritime environments, several key design strategies should be considered. First, integrating temporal memory and re-identification modules is essential, as maritime tracking often involves long-term occlusions and the re-emergence of targets. In addition, using multi-modal inputs such as combining RGB data with radar or depth sensing can significantly reduce ambiguity in scenarios with poor visibility or cluttered backgrounds. It is also important to integrate semantic knowledge and context modeling; for instance, using domain knowledge such as vessel hull shapes and motion dynamics can improve robustness under visually complex conditions. In addition, training pipelines should be equipped with maritime-specific data augmentations, simulating phenomena such as glare, wave-induced distortions, and partial occlusions to increase robustness. Finally, benchmarking frameworks should support evaluations under both generalization and adaptation protocols, as demonstrated in our Protocols I and II analyses, to accurately assess tracker performance.
\vspace{-4.5mm}
\subsection{Key Insights and Implications for Future Research}

The evaluation of trackers on MVTD provides several important takeaways for the tracking research community. First and foremost, maritime tracking is fundamentally different from general-purpose object tracking. The domain-specific visual and dynamic conditions—such as fluctuating lighting, reflective water surfaces, and frequent occlusions—make it inadequate to rely solely on open-air datasets for performance benchmarking. Second, our results strongly confirm that fine-tuning on domain-specific data significantly enhances tracker robustness. Even a modest amount of maritime-specific training data can yield substantial improvements in accuracy, underscoring the critical role of transfer learning for real-world deployment. Third, there is no one-size-fits-all solution; tracker performance varies drastically based on architectural design and the particular maritime challenges presented. Therefore, tracker selection should be context-aware, tailored to operational environments such as port surveillance, offshore monitoring, or autonomous navigation. Lastly, the proposed MVTD dataset addresses a critical gap in the current landscape. With its inclusion of maritime-relevant attributes, both long- and short-term sequences, and diverse camera perspectives, MVTD serves as a high-fidelity testbed for evaluating and developing robust tracking algorithms tailored to the complexities of maritime environments.

\vspace{-5mm}
\section{Conclusion}\label{sec:conclusion}

In this work, we presented Maritime Visual Tracking Dataset (MVTD), a novel and comprehensive visual object tracking dataset specifically designed for maritime environments. The dataset encompasses 182 high-resolution video sequences with over 150,000 manually annotated frames, covering diverse maritime objects such as boats, ship, sailboats, and USV. MVTD introduces a wide array of domain-specific challenges including water surface reflections, varying illumination conditions, scale changes, partial visibility, and low-contrast backgrounds, which are rarely encountered or insufficiently represented in open-air tracking datasets.

We conducted extensive evaluations across two protocols: (i) inference using pre-trained SOTA trackers, and (ii) fine-tuning and re-evaluation on our dataset. Results from 14 advanced tracking algorithms reveals that trackers experience significant performance degradation when applied to maritime conditions. However, fine-tuning on MVTD led to substantial gains in success rate, precision, and normalized precision, underlining the critical importance of domain adaptation in tracking application.

The proposed dataset, along with our evaluation protocols and baseline results, sets a new benchmark for maritime object tracking. It enables researchers to systematically study the limitations of current trackers and motivates the development of more robust algorithms capable of addressing the inherent challenges posed by maritime environments. MVTD fills a critical gap in the existing literature by providing both the data and benchmarking tools necessary to drive forward innovation in real-world tracking applications such as maritime surveillance, coastal monitoring, and autonomous navigation.
\noindent
\balance

\bibliographystyle{IEEEtran}
\bibliography{references}

\end{document}